\documentclass[11pt,a4paper]{article}
\usepackage{authblk} 
\usepackage[hyperref]{acl2017}
\usepackage{times}
\usepackage{url}
\usepackage{latexsym}
\usepackage[T1]{fontenc}    % use 8-bit T1 fonts
\usepackage{hyperref}       % hyperlinks
\usepackage{url}            % simple URL typesetting
\usepackage{booktabs}       % professional-quality tables
\usepackage{amsfonts}       % blackboard math symbols
\usepackage{nicefrac}       % compact symbols for 1/2, etc.
\usepackage{microtype}      % microtypography
\usepackage{amsmath}
\usepackage{mathtools}
\usepackage{graphicx,color,xcolor}
\usepackage{multirow}
\usepackage{enumitem} 
\usepackage{bm}
\usepackage{algorithm}
\usepackage[noend]{algpseudocode}

\usepackage{todonotes}

\graphicspath{{./figures}}

\usepackage{lipsum}

\newcommand\blfootnote[1]{%
  \begingroup
  \renewcommand\thefootnote{}\footnote{#1}%
  \addtocounter{footnote}{-1}%
  \endgroup
}

\DeclareMathOperator*{\argmax}{arg\,max}

\aclfinalcopy % Uncomment this line for the final submission
\def\aclpaperid{177} %  Enter the acl Paper ID here
\makeatletter
\def\BState{\State\hskip-\ALG@thistlm}
\makeatother
\title{Trainable Greedy Decoding for Neural Machine Translation}
\def \nyu{$^\ddag$}
\def \hku{$^\dagger$}

\author[\hku]{\bf Jiatao Gu}
\author[\nyu]{\bf Kyunghyun Cho}
\author[\hku]{\bf Victor O.K. Li}
\affil[\hku]{The University of Hong Kong}
\affil[\nyu]{New York University}
\affil[\hku]{\tt  \{jiataogu, vli\}@eee.hku.hk}
\affil[\nyu]{\tt  kyunghyun.cho@nyu.edu}

\date{}

\begin{document}
\maketitle

\begin{abstract}

Recent research in neural machine translation has largely focused on two aspects; neural network architectures and end-to-end learning algorithms. The problem of decoding, however, has received relatively little attention from the research community. In this paper, we solely focus on the problem of decoding given a trained neural machine translation model. Instead of trying to build a new decoding algorithm for any specific decoding objective, we propose the idea of {\it trainable decoding algorithm} in which we train a decoding algorithm to find a translation that maximizes an arbitrary decoding objective. More specifically, we design an actor that observes and manipulates the hidden state of the neural machine translation decoder and propose to train it using a variant of deterministic policy gradient. We extensively evaluate the proposed algorithm using four language pairs and two decoding objectives, and show that we can indeed train a trainable greedy decoder that generates a better translation (in terms of a target decoding objective) with minimal computational overhead.\blfootnote{The majority of this work was completed while the first author was visiting New York University.} 

\end{abstract}

\section{Introduction}
\label{sec:introduction}

Neural machine translation has recently become a method of choice in machine translation research. Besides its success in traditional settings of machine translation, that is one-to-one translation between two languages, \citep{sennrich2016edinburgh,chung2016nyu}, neural machine translation has ventured into more sophisticated settings of machine translation. For instance, neural machine translation has successfully proven itself to be capable of handling subword-level representation of sentences \citep{lee2016fully,luong2016achieving,sennrich2015neural,costa2016character,ling2015character}. Furthermore, several research groups have shown its potential in seamlessly handling multiple languages \citep{dong2015multi,luong2015multi,firat2016multi,firat2016zero,lee2016fully,ha2016toward,viegas2016google}. 

A typical scenario of neural machine translation starts with training a model to maximize its log-likelihood. That is, we often train a model to maximize the conditional probability of a reference translation given a source sentence over a large parallel corpus. Once the model is trained in this way, it defines the conditional distribution over all possible translations given a source sentence, and the task of translation becomes equivalent to finding a translation to which the model assigns the highest conditional probability. Since it is computationally intractable to do so exactly, it is a usual practice to resort to approximate search/decoding algorithms such as greedy decoding or beam search. In this scenario, we have identified two points where improvements could be made. They are (1) training (including the selection of a model architecture) and (2) decoding.

% \begin{figure}[t]
% \centering
% \includegraphics[width=0.85\linewidth]{figures/Concept.pdf}
% \caption{\label{fig:framework} The graphical illustration of the proposal trainable greedy decoding. $\theta$, $\phi$ and $\psi$ respectively correspond to the parameters of the underlying neural machine translation model, the trainable greedy decoding (actor) and the critic. 
% }
% \end{figure}

Much of the research on neural machine translation has focused solely on the former, that is, on improving the model architecture. Neural machine translation started with with a simple encoder-decoder architecture in which a source sentence is encoded into a single, fixed-size vector \citep{cho2014learning,sutskever2014sequence,kalchbrenner2013recurrent}. 
It soon evolved with the attention mechanism \citep{bahdanau2014neural}. 
A few variants of the attention mechanism, or its regularization, have been proposed recently to improve both the translation quality as well as the computational efficiency \citep{luong2015effective,cohn2016incorporating,tu2016modeling}. 
More recently, convolutional networks have been adopted either as a replacement of or a complement to a recurrent network in order to efficiently utilize parallel computing 
\citep{kalchbrenner2016neural,lee2016fully,gehring2016convolutional}.

On the aspect of decoding, only a few research groups have tackled this problem by incorporating a target decoding algorithm into training. \citet{wiseman2016sequence} and \citet{shen2015minimum} proposed a learning algorithm tailored for beam search. \citet{ranzato2015sequence} and \cite{bahdanau2016actor} suggested to use a reinforcement learning algorithm by viewing a neural machine translation model as a policy function.
Investigation on decoding alone has, however, been limited. \citet{cho2016noisy} showed the limitation of greedy decoding by simply injecting unstructured noise into the hidden state of the neural machine translation system. \citet{tu2016neural} similarly showed that the exactness of beam search does not correlate well with actual translation quality, and proposed to augment the learning cost function with reconstruction to alleviate this problem. \citet{li2016simple} proposed a modification to the existing beam search algorithm to improve its exploration of the translation space. 

In this paper, we tackle the problem of decoding in neural machine translation by introducing a concept of {\it trainable greedy decoding}. Instead of manually designing a new decoding algorithm suitable for neural machine translation, we propose to learn a decoding algorithm with an arbitrary decoding objective. More specifically, we introduce a neural-network-based decoding algorithm that works on an already-trained neural machine translation system by observing and manipulating its hidden state. We treat such a neural network as an agent with a deterministic, continuous action and train it with a variant of the deterministic policy gradient algorithm \citep{silver2014deterministic}. 

We extensively evaluate the proposed trainable greedy decoding on four language pairs (En-Cs, En-De, En-Ru and En-Fi; in both directions) with two different decoding objectives; sentence-level BLEU and negative perplexity. By training such trainable greedy decoding using deterministic policy gradient with the proposed critic-aware actor learning, we observe that we can improve decoding performance with minimal computational overhead. Furthermore, the trained actors are found to improve beam search as well, suggesting a future research direction in extending the proposed idea of trainable decoding for more sophisticated underlying decoding algorithms.

\section{Background}

\subsection{Neural Machine Translation}

Neural machine translation is a special case of conditional recurrent language modeling, where the source and target are natural language sentences. Let us use $X=\left\{ x_1, \ldots, x_{T_s} \right\}$ and $Y=\left\{ y_1, \ldots, y_T \right\}$ to denote source and target sentences, respectively. Neural machine translation then models the target sentence given the source sentence as:
%\begin{align*}
$p(Y|X) = \prod_{t=1}^T p(y_t | y_{<t}, X)$.
%\end{align*}
Each term on the r.h.s. of the equation above is modelled as a composite of two parametric functions:
\begin{align*}
p(y_t|y_{<t}, X)\propto \exp\left(g\left(y_t, z_t; \theta_g\right)\right),
\end{align*}
where 
%\begin{align*}
$z_t = f(z_{t-1}, y_{t-1}, e_t(X; \theta_e); \theta_f)$.
%\end{align*}
$g$ is a read-out function that transforms the hidden state $z_t$ into the distribution over all possible symbols, and $f$ is a recurrent function that compresses all the previous target words $y_{<t}$ and the time-dependent representation $e_t(X; \theta_e)$ of the source sentence $X$. This time-dependent representation $e_t$ is often implemented as a recurrent network encoder of the source sentence coupled with an attention mechanism \citep{bahdanau2014neural}.

% Neural machine translation~\cite{sutskever2014sequence,cho2014learning,bahdanau2014neural} has recently achieved impressive improvement~\cite{wu2016google} compared to traditional statistical machine translation methods. Across different variants, the NMT model typically models the machine translation as a auto-regressive generative model. That is to say, given a source sequence $X=\{x_1, ..., x_{Ts}\}$, the distribution of a translation sequence $Y = \{y_1, ..., y_T\}$ can be computed as:
% \begin{equation}
% \label{eq.crnnlm}
% p(Y|X) = \prod_{t=1}^T p(y_t|y_{<t}, X)
% \end{equation}
% where the conditional probability is modeled as the composite of two parametric functions:
% \begin{equation}
% \label{eq.model}
% \begin{split}
% &p(y_t|y_{<t}, X)\propto \exp\left(g\left(y_t, z_t; \theta_g\right)\right) \\
% &z_t = f(z_{t-1}, y_{t-1}, e_t(X; \theta_e); \theta_f)
% \end{split}
% \end{equation}
% where $e_t(X; \theta_e)$ is a time-dependent feature of the source sentence $X$ extracted by, for instance, a bi-directional recurrent neural network (RNN) with attention mechanism~\cite{bahdanau2014neural}, or so-called an \textit{encoder}; $f$ is a \textit{decoder} which is usually modeled by a separate RNN and $z_t$ is the hidden states of the decoder at step $t$; $g$ is the energy function which maps the hidden states into a distribution over the vocabulary. $\theta_g$, $\theta_f$ and $\theta_e$ are the parameters for $g$, $f$ and the encoder respectively.

\paragraph{Maximum Likelihood Learning}

We train a neural machine translation model, or equivalently estimate the parameters $\theta_g$, $\theta_f$ and $\theta_e$, by maximizing the log-probability of a reference translation $\hat{Y}=\{\hat{y}_1, ..., \hat{y}_T\}$ given a source sentence. That is, we maximize the log-likelihood function:
\begin{align*}
J^{\text{ML}}(\theta_g, \theta_f, \theta_e) = \frac{1}{N} \sum_{n=1}^N \sum_{t=1}^{T_n} \log p_{\theta}(\hat{y}_t^n| \hat{y}_{<t}^n, X^n),
\end{align*}
given a training set consisting of $N$ source-target sentence pairs. It is important to note that this maximum likelihood learning does not take into account how a trained model would be used. Rather, it is only concerned with learning a distribution over all possible translations.

\subsection{Decoding}

Once the model is trained, either by maximum likelihood learning or by any other recently proposed algorithms \citep{wiseman2016sequence,shen2015minimum,bahdanau2016actor,ranzato2015sequence}, we can let the model translate a given sentence by finding a translation that maximizes 
\begin{align*}
\hat{Y} = \argmax_{Y} \log p_{\theta} (Y|X),
\end{align*}
where $\theta=(\theta_g, \theta_f, \theta_e)$.
This is, however, computationally intractable, and it is a usual practice to resort to approximate decoding algorithms.

\paragraph{Greedy Decoding}

One such approximate decoding algorithm is greedy decoding. In greedy decoding, we follow the conditional dependency path and pick the symbol with the highest conditional probability so far at each node. This is equivalent to picking the best symbol one at a time from left to right in conditional language modelling. A decoded translation of greedy decoding is $\hat{Y} = (\hat{y}_1, \ldots, \hat{y}_T)$, where
\begin{equation}
\hat{y}_t =  \argmax_{y \in V} \log p_{\theta}(y|\hat{y}_{<t}, X).
\end{equation}
Despite its preferable computational complexity $O(|V| \times T)$, greedy decoding has been over time found to be undesirably sub-optimal.% (see, e.g., \citep{cho2016noisy}.) 

\paragraph{Beam Search}  

Beam search keeps $K > 1$ hypotheses, unlike greedy decoding which keeps only a single hypothesis during decoding. At each time step $t$, beam search picks $K$ hypotheses with the highest scores ($\prod_{t'=1}^t p(y_t | y_{<t}, X)$). When all the hypotheses terminate (outputting the end-of-the-sentence symbol), it returns the hypothesis with the highest log-probability. Despite its superior performance compared to greedy decoding, the computational complexity grows linearly w.r.t. the size of beam $K$, which makes it less preferable especially in the production environment.

% \paragraph{\textbf{Training \& Testing Mismatch}} There is clearly a mismatch between the conventional pipeline. The NMT model is trained as a stochastic 

% \paragraph{\textbf{Metric Mismatch}} The target of maximum likelihood mismatches the final sequence-level evaluation metrics, for instance BLEU score in machine translation. That is to say, even if we can find a good decoding trajectory based on the trained conditional language model $\theta$, it still does not guarantee to obtain the best BLEU score in the end. 

\section{Trainable Greedy Decoding}

\subsection{Many Decoding Objectives}

Although we have described decoding in neural machine translation as a maximum-a-posteriori estimation in $\log p(Y|X)$, this is not necessarily the only %decoding objective 
nor the desirable decoding objective. 

First, each potential scenario in which neural machine translation is used calls for a unique decoding objective. In simultaneous translation/interpretation, which has recently been studied in the context of neural machine translation \citep{gu2016learning}, the decoding objective is formulated as a trade-off between the translation quality and delay. On the other hand, when a machine translation system is used as a part of a larger information extraction system, it is more important to correctly translate named entities and events than to translate syntactic function words. The decoding objective in this case must account for how the translation is used in subsequent modules in a larger system. 

Second, the conditional probability assigned by a trained neural machine translation model does not necessarily reflect our perception of translation quality. Although \citet{cho2016noisy} provided empirical evidence of high correlation between the log-probability and BLEU, a {\it de facto} standard metric in machine translation, there have also been reports on large mismatch between the log-probability and BLEU. For instance, \citet{tu2016neural} showed that beam search with a very large beam, which is supposed to find translations with better log-probabilities, suffers from pathological translations of very short length, resulting in low translation quality. This calls for a way to design or {\it learn} a decoding algorithm with an objective that is more directly correlated to translation quality. 

In short, there is a significant need for designing multiple decoding algorithms for neural machine translation, regardless of how it was trained. It is however non-trivial to manually design a new decoding algorithm with an arbitrary objective. This is especially true with neural machine translation, as the underlying structure of the decoding/search process -- the high-dimensional hidden state of a recurrent network -- is accessible but not interpretable. Instead, in the remainder of this section, we propose our approach of {\it trainable greedy decoding}.

\begin{figure*}[t]
\centering
\begin{minipage}{0.69\textwidth}
\includegraphics[width=\columnwidth]{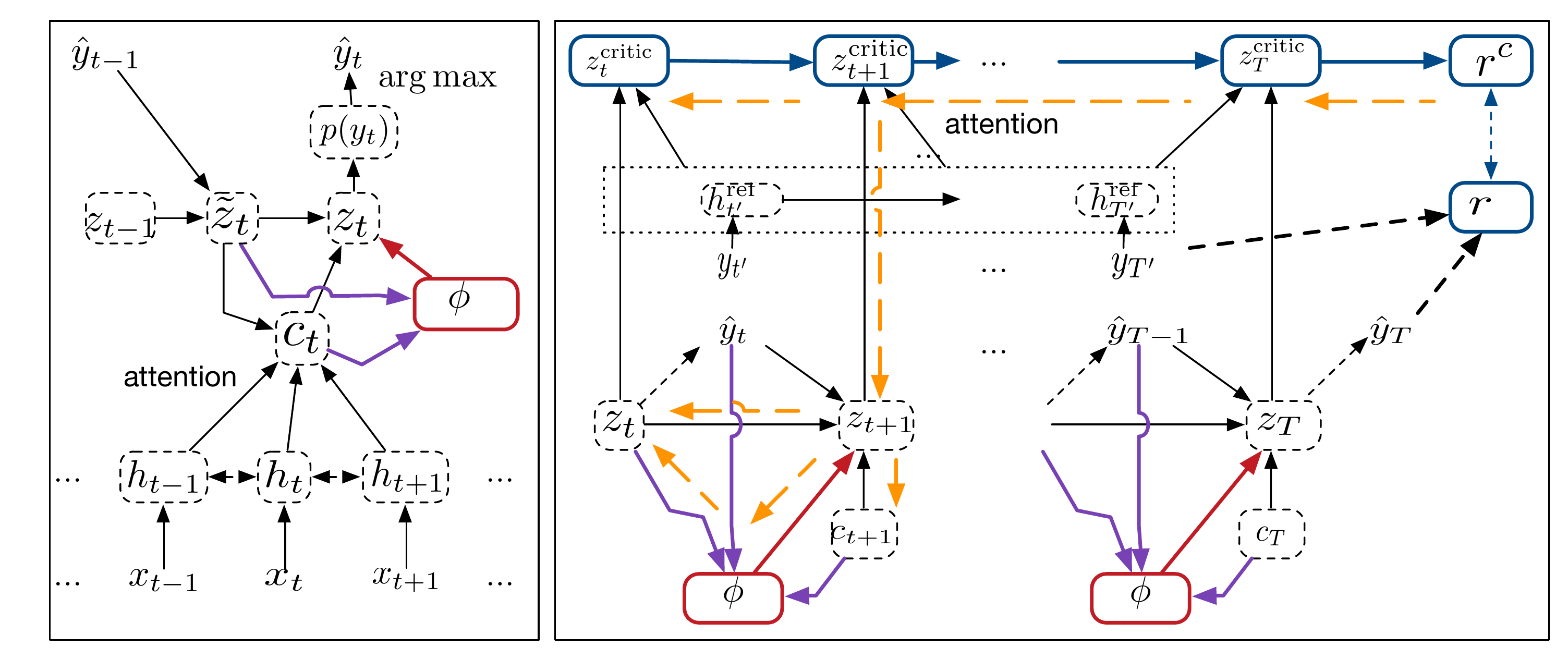}
\end{minipage}
\hfill
\begin{minipage}{0.30\textwidth}
\caption{\label{fig:tgd}  
\small Graphical illustrations of the trainable greedy decoding. The left panel shows a single step of the actor interacting with the underlying neural translation model, and The right panel the interaction among the underlying neural translation system (dashed-border boxes), actor (red-border boxes), and critic (blue-border boxes). The solid arrows indicate the forward pass, and the dashed yellow arrows the actor's backward pass. The dotted-border box shows the use of a reference translation.}
\end{minipage}

\vspace{-4mm}
\end{figure*}

\subsection{Trainable Greedy Decoding}

We start from the noisy, parallel approximate decoding (NPAD) algorithm proposed in \citep{cho2016noisy}. The main idea behind NPAD algorithm is that a better translation with a higher log-probability may be found by injecting unstructured noise in the transition function of a recurrent network. That is,
\begin{align*}
z_t = f(z_{t-1} + \epsilon_t, y_{t-1}, e_t(X; \theta_e); \theta_f),
\end{align*}
where $\epsilon_t \sim \mathcal{N}(0, (\sigma_0/t)^2)$. NPAD avoids potential degradation of translation quality by running such a noisy greedy decoding process multiple times in parallel. An important lesson of NPAD algorithm is that there exists a decoding strategy with the asymptotically same computational complexity that results in a better translation quality, and that such a better translation can be found by manipulating the hidden state of the recurrent network. 

In this work, we propose to significantly extend NPAD by replacing the unstructured noise $\epsilon_t$ with a parametric function approximator, or an agent, $\pi_{\phi}$. This agent takes as input the previous hidden state $z_{t-1}$, previously decoded word $\hat{y}_{t-1}$ and the time-dependent context vector $e_t(X; \theta_e)$ and outputs a real-valued vectorial action $a_t \in \mathbb{R}^{\text{dim}(z_t)}$. Such an agent is trained such that greedy decoding with the agent finds a translation that maximizes any predefined, arbitrary decoding objective, while the underlying neural machine translation model is pretrained and fixed. Once the agent is trained, we generate a translation given a source sentence by greedy decoding however augmented with this agent. We call this decoding strategy {\it trainable greedy decoding}. 

\paragraph{Related Work: Soothsayer prediction function}

Independently from and concurrently with our work here, \citet{li2017learning} proposed, just two weeks earlier, to train a neural network that predicts an arbitrary decoding objective given a source sentence and a partial hypothesis, or a prefix of translation, and to use it as an auxiliary score in beam search. For training such a network, referred to as a Q network in their paper, they generate each training example by either running beam search or using a ground-truth translation (when appropriate) for each source sentence. This approach allows one to use an arbitrary decoding objective, but it still relies heavily on the log-probability of the underlying neural translation system in actual decoding. We expect a combination of these and our approaches may further improve decoding for neural machine translation in the future.

\subsection{Learning and Challenges}

While all the parameters---$\theta_g$, $\theta_f$ and $\theta_e$--- of the underlying neural translation model are fixed, we only update the parameters $\phi$ of the agent $\pi$. This ensures the generality of the pretrained translation model, and allows us to train multiple trainable greedy decoding agents with different decoding objectives, maximizing the utility of a single trained translation model. 

Let us denote by $R$ our arbitrary decoding objective as a function that scores a translation generated from trainable greedy decoding. Then, our learning objective for trainable greedy decoding is 
\begin{align*}
J^{\text{A}}(\phi) = \mathbb{E}_{X \sim D}^{\hat{Y}=G_{\pi}(X)}\left[R(\hat{Y})\right],
\end{align*}
where we used $G_{\pi}(X)$ as a shorthand for trainable greedy decoding with an agent $\pi$. 

There are two major challenges in learning an agent with such an objective. First, the decoding objective $R$ may not be differentiable with respect to the agent. Especially because our goal is to accommodate an arbitrary decoding objective, this becomes a problem. For instance, BLEU, a standard quality metric in machine translation, is a piece-wise linear function with zero derivatives almost everywhere. Second, the agent here is a real-valued, deterministic policy with a very high-dimensional action space (1000s of dimensions), which is well known to be difficult. In order to alleviate these difficulties, we propose to use a variant of the deterministic policy gradient algorithm \citep{silver2014deterministic,lillicrap2015continuous}.

\section{Deterministic Policy Gradient \\ with Critic-Aware Actor Learning}

\subsection{Deterministic Policy Gradient \\ for Trainable Greedy Decoding}

It is highly unlikely for us to have access to the gradient of an arbitrary decoding objective $R$ with respect to the agent $\pi$, or its parameters $\phi$. Furthermore, we cannot estimate it stochastically because our policy $\pi$ is defined to be deterministic without a predefined nor learned distribution over the action. Instead, following \citep{silver2014deterministic,lillicrap2015continuous}, we use a parametric, differentiable approximator, called a critic $R^c$, for the non-differentiable objective $R$. We train the critic by minimizing
\begin{align*}
J^{\text{C}}(\psi) = \mathbb{E}_{X \sim D}^{\hat{Y}=G_{\pi}(X)}\left[R^c_{\psi}(z_{1:T}) - R(\hat{Y})\right]^2.
\end{align*}
The critic observes the state-action sequence of the agent $\pi$ via the modified hidden states $(z_1, \ldots, z_T)$ of the recurrent network, and predicts the associated decoding objective. By minimizing the mean squared error above, we effectively encourage the critic to approximate the non-differentiable objective as closely as possible in the vicinity of the state-action sequence visited by the agent. 

We implement the critic $R^c$ as a recurrent network, similarly to the underlying neural machine translation system. This implies that we can compute the derivative of the predicted decoding objective with respect to the input, that is, the state-action sequence $z_{1:T}$, which allows us to update the actor $\pi$, or equivalently its parameters $\phi$, to maximize the predicted decoding objective. Effectively we avoid the issue of non-differentiability of the original decoding objective by working with its proxy. 

With the critic, the learning objective of the actor is now to maximize not the original decoding objective $R$ but its proxy $R^{\text{C}}$ such that
\begin{align*}
\hat{J}^{\text{A}}(\phi) = \mathbb{E}_{X \sim D}^{\hat{Y}=G_{\pi}(X)}\left[R^{\text{C}}(\hat{Y})\right].
\end{align*}
Unlike the original objective, this objective function is fully differentiable with respect to the agent $\pi$. We thus use a usual stochastic gradient descent algorithm to train the agent, while simultaneously training the critic. We do so by alternating between training the actor and critic. Note that we maximize the return of a full episode rather than the Q value, unlike usual approaches in reinforcement learning.

\begin{algorithm}[t]
\caption{Trainable Greedy Decoding}
\label{algo2}
\begin{algorithmic}[1]
\small
\Require{NMT $\theta$, actor $\phi$, critic $\psi$, $N_c$, $N_a$, $S_c$, $S_a$, $\tau$}
\State Train $\theta$ using MLE on training set $D$;
\State Initialize $\phi$ and $\psi$;
\State Shuffle $D$ twice into $D_{\phi}$ and $D_{\psi}$
\While{stopping criterion is not met}
\For{$t=1:N_c$}
\State Draw a translation pair: $(X, Y)\sim D_{\psi}$;
\State $r, r^c=\textsc{Decode}(S_c, X, Y, 1)$
\State Update $\psi$ using $\nabla_{\psi}\sum_k{\left(r_k^c - r_k\right)^2}/(S_c+1)$
\EndFor
\For{$t=1:N_a$}
\State Draw a translation pair: $(X, Y)\sim D_{\phi}$;
\State $r, r^c=\textsc{Decode}(S_a, X, Y, 0)$
\State Compute $w_k = \exp\left(-\left(r_k^c - r_k\right)^2/\tau\right)$
\State Compute $\tilde{w}_k=w_k/\sum_k{w_k}$
\State Update $\phi$ using $-\sum_k{\left(\tilde{w}_k\cdot \nabla_{\phi}r^c_k\right)}$
\EndFor
\EndWhile
\Statex{}
\setcounter{ALG@line}{0}
\Statex{\hspace{-18pt}\textbf{Function: }}{\textsc{Decode}$(S, X, Y, c)$}
  \State $Y_s = \{\}$, $Z_s = \{\}$, $r = \{\}$, $r^c=\{\}$;
  \For{$k=1:S$}
   \State Sample noise $\epsilon \sim \mathcal{N}(0, \sigma^2)$ for each action;
   \State Greedy decoding $\hat{Y}^{k} = G_{\theta, \phi}(X)$ with $\epsilon$;
   \State Collect hidden states $z^{k}_{1:T}$ given $X$, $\hat{Y}$, $\theta$, $\phi$
   \State $Y_s \leftarrow Y_s \cup \{Y^k\}$
   \State $Z_s \leftarrow Z_s \cup \{z^{k}_{1:T}\}$
  \EndFor
  \If{$c=1$}
   \State Collect hidden states $z_{1:T}$ given $X$, $Y$, $\theta$
   \State $Y_s \leftarrow Y_s \cup \{Y\}$
   \State $Z_s \leftarrow Z_s \cup \{ z_{1:T} \}$
  \EndIf
  \For{$\hat{Y}, Z \in Y_s, Z_s$} 
   \State Compute the critic output $r^c \leftarrow R^c_{\psi}(Z, \hat{Y})$ 
   \State Compute true reward $r \leftarrow R(Y, \hat{Y})$
  \EndFor
  \State \textbf{return} $r$, $r^c$
\end{algorithmic}
\end{algorithm}

\begin{figure*}[t]
\centering
\begin{minipage}{0.71\textwidth}
\centering

\begin{minipage}{0.48\columnwidth}
\centering
\includegraphics[width=\columnwidth]{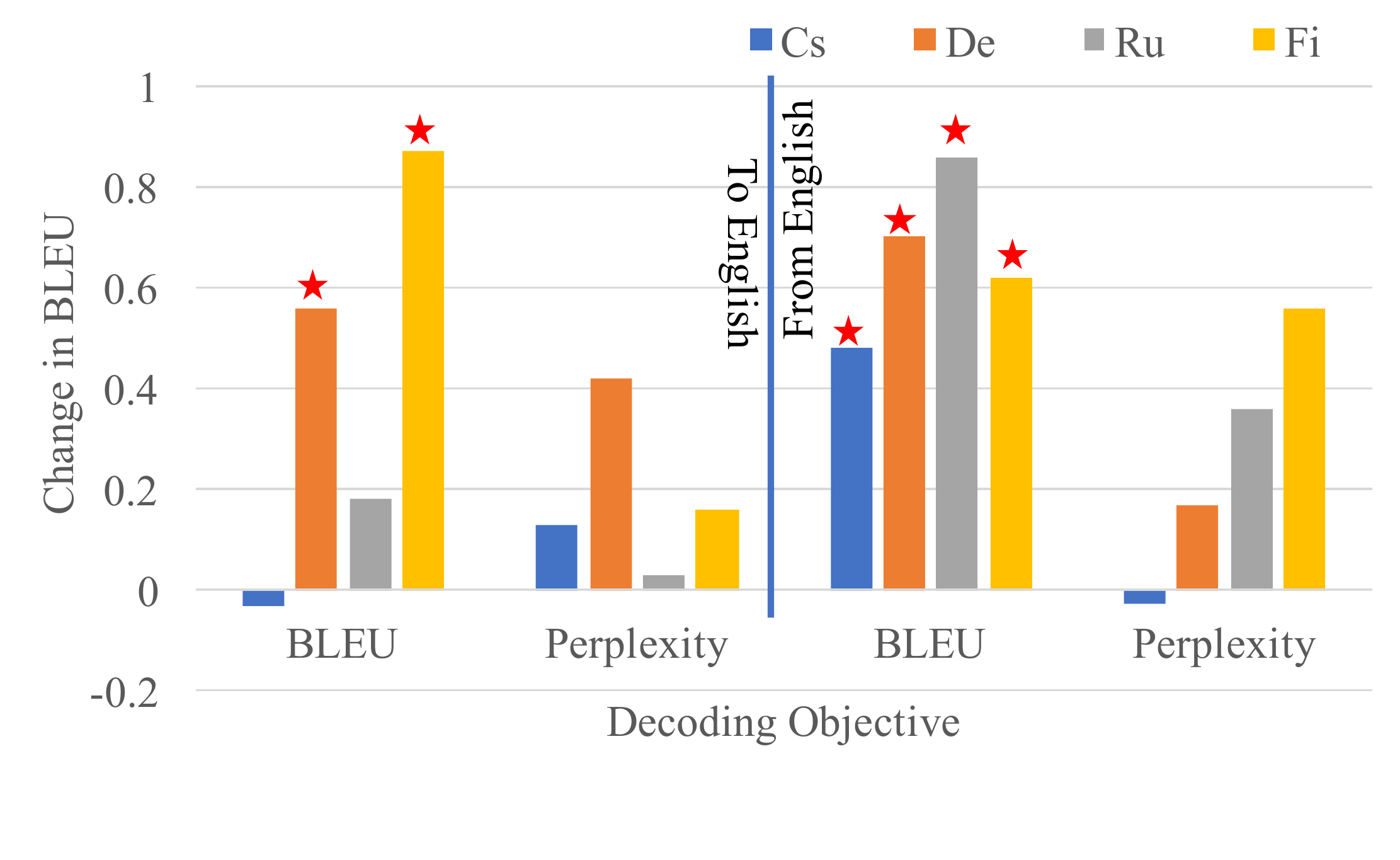}
\end{minipage}
\hfill
\begin{minipage}{0.48\columnwidth}
\centering
\includegraphics[width=\columnwidth]{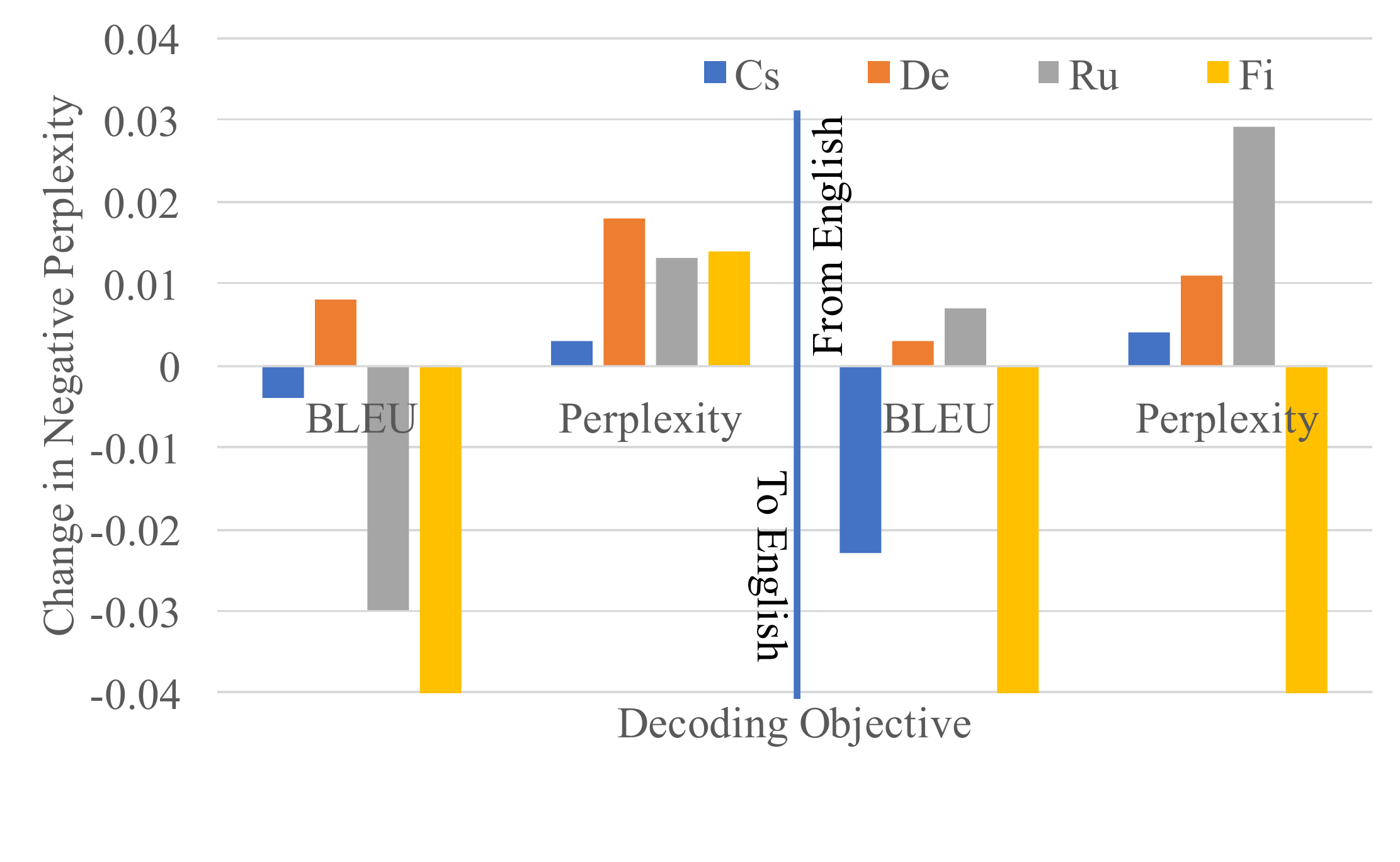}
\end{minipage}
\label{fig:r1}

\vspace{-3mm}
(a) Trainable Greedy Decoding

\begin{minipage}{0.48\columnwidth}
\centering
\includegraphics[width=\columnwidth]{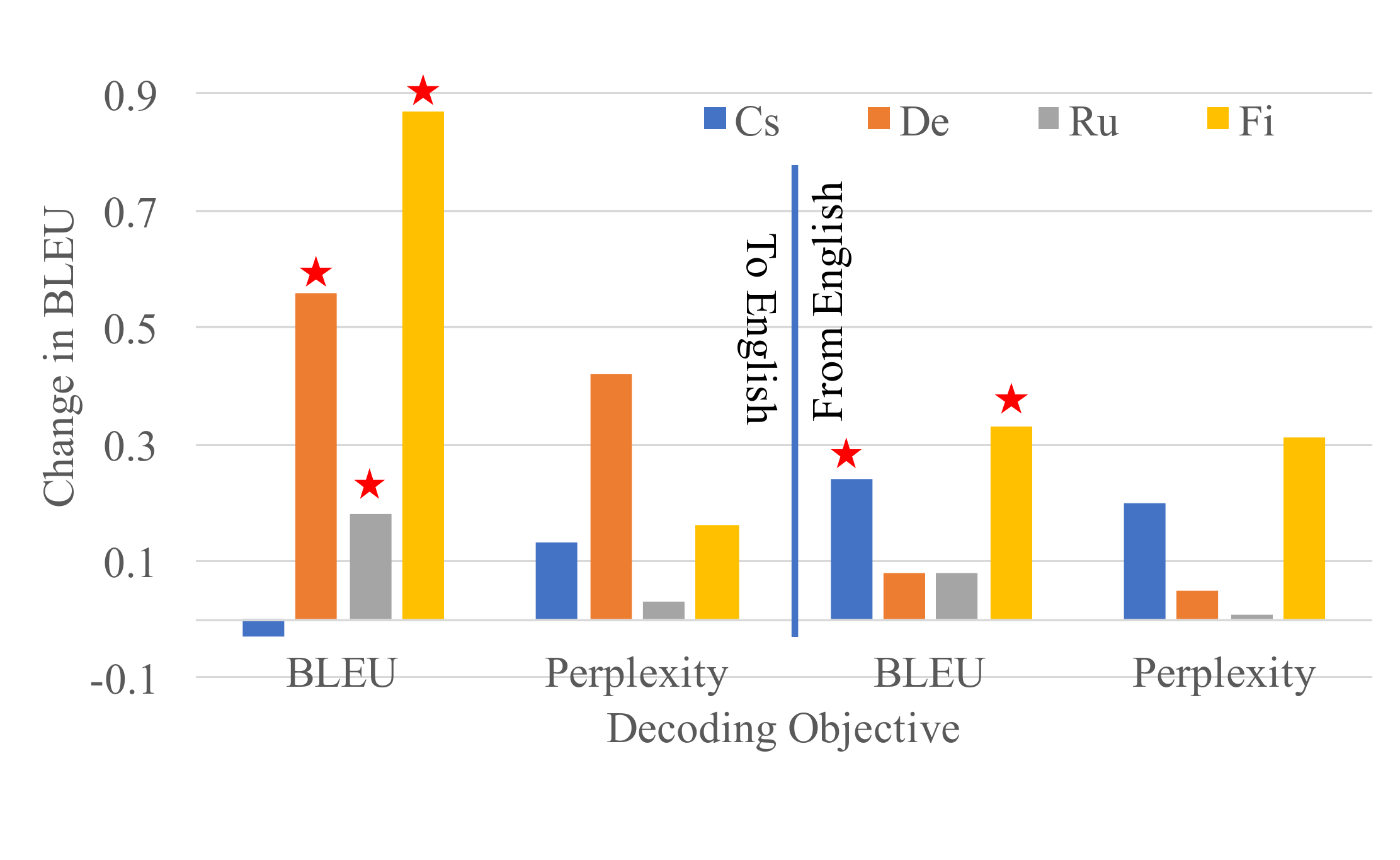}
\end{minipage}
\hfill
\begin{minipage}{0.48\columnwidth}
\centering
\includegraphics[width=\columnwidth]{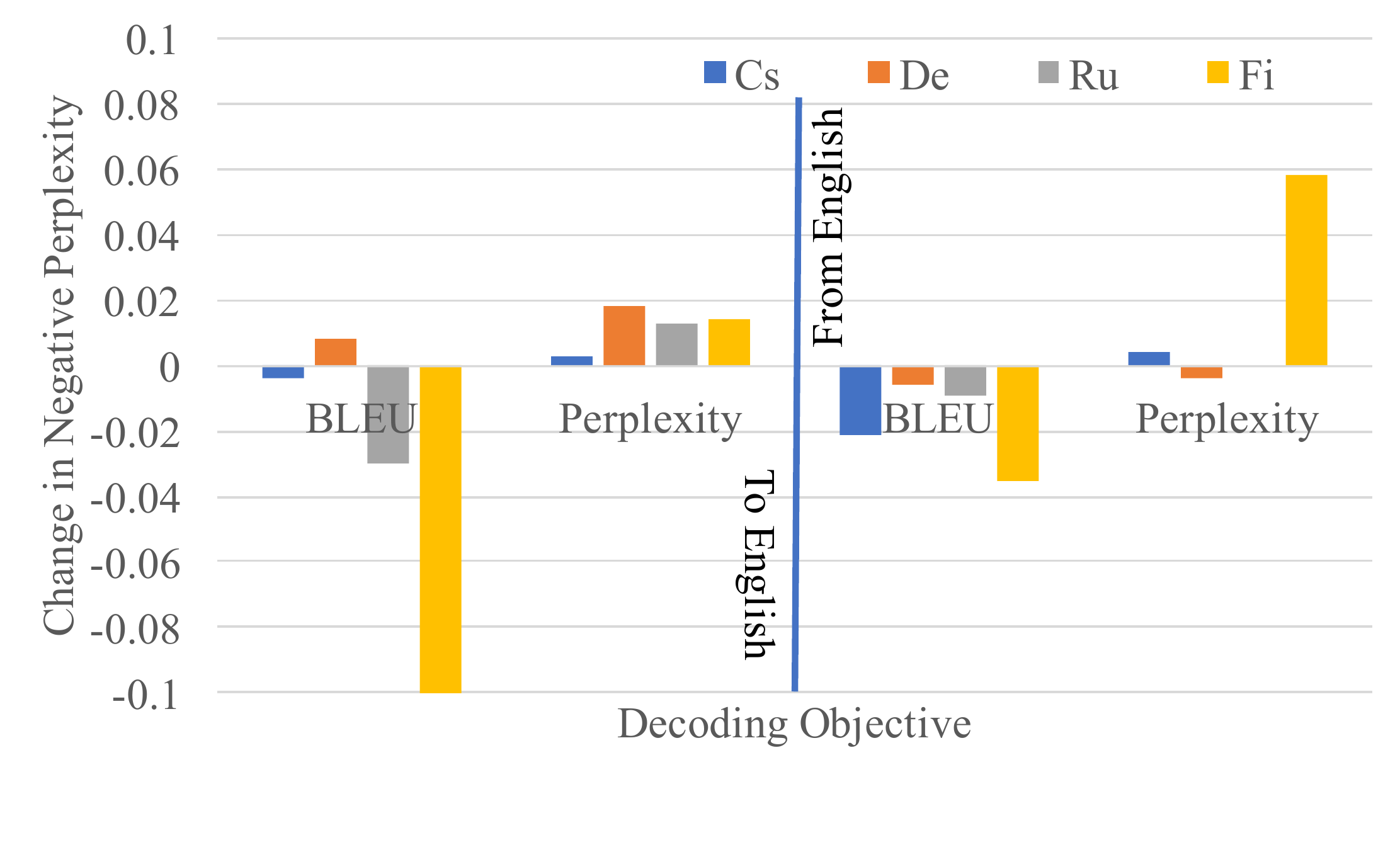}
\end{minipage}
\label{fig:r2}

\vspace{-3mm}
(b) Beam Search + Trainable Greedy Decoding
\end{minipage}
\hfill
\begin{minipage}{0.28\textwidth}
\caption{
\label{fig:result1} \small
The plots draw the improvements by the trainable greedy decoding. The x-axes correspond to the objectives used to train trainable greedy decoding, and the y-axes to the changes in the achieved objectives (BLEU for the figures on the left, and negative perplexity on the right.) The top row (a) shows the cases when the trainable greedy decoder is used on its own, and the bottom row (b) when it is used together with beam search. The baselines (improvement of 0) are respectively greedy decoding and beam search for the top and bottom rows. When training and evaluation are both done with BLEU, we test the statistical significance \citep{koehn2004statistical}, and we mark significant cases with red stars ($p < 0.05$.) 
}
\end{minipage}
\vspace{-4mm}
\end{figure*}

\subsection{Critic-Aware Actor Learning}

\paragraph{Challenges}

The most apparent challenge for training such a deterministic actor with a large action space is that most of action configurations will lead to zero return. It is also not trivial to devise an efficient exploration strategy with a deterministic actor with real-valued actions. This issue has however turned out to be less of a problem than in a usual reinforcement learning setting, as the state and action spaces are well structured thanks to pretraining by maximum likelihood learning. As observed by \citet{cho2016noisy}, any reasonable perturbation to the hidden state of the recurrent network generates a reasonable translation which would receive again a reasonable return. 

Although this property of dense reward makes the problem of trainable greedy decoding more manageable, we have observed other issues during our preliminary experiment with the vanilla deterministic policy gradient. In order to avoid these issues that caused instability, we propose the following modifications to the vanilla algorithm.

% \label{sec.stabilize}
% When applying the proposed trainable decoding algorithm into practice, there is one notable property. That is, the action space is unprecedentedly high-dimensional, as the proposed actions are additive to the hidden states. Most of NMT models often have a 1000s, if not 10s of 1000s, dimensional hidden units in the decoder, making it an extremely challenging reinforcement learning problem. 
% This problem is however lessened significantly in our settings as the maximum likelihood training shapes a well-structured space for the decoder's hidden states and the actor exploits the hidden states by injecting actions on them. This constrains the exploration, the biggest challenge of learning in high-dimensional continuous space, substantially easier.

% In addition, we also proposed several methods to further stabilize the learning procedure.

\vspace{-1mm}
\paragraph{Critic-Aware Actor Learning}
A major goal of the critic is not to estimate the return of a given episode, but to estimate the gradient of the return evaluated given an episode. In order to do so, the critic must be trained, or presented, with state-action sequences $z_{1:T'}$ similar though not identical to the state-action sequence generated by the current actor $\pi$. This is achieved, in our case, by injecting unstructured noise to the action at each time step, similar to \citep{heess2015learning}:
\begin{align}
\vspace{-5pt}
\label{eq:noisy_actor}
\tilde{a}_t = \phi(z_t, a_{t-1}) + \sigma \cdot \epsilon,
\end{align}
where $\epsilon$ is a zero-mean, unit-variance normal variable. This noise injection procedure is mainly used when training the critic. 

We have however observed that the quality of the reward and its gradient estimate of the critic is very noisy even when the critic was trained with this kind of noisy actor. This imperfection of the critic often led to the instability in training the actor in our preliminary experiments. In order to avoid this, we describe here a technique which we refer to as {\it critic-aware actor gradient estimation}.

Instead of using the point estimate $\frac{\partial R^c}{\partial \phi}$ of the gradient of the predicted objective with respect to the actor's parameters $\phi$, we propose to use the expected gradient of the predicted objective with respect to the critic-aware distribution $Q$. That is,
\begin{align}
\label{eq:critic-aware}
\mathbb{E}_{Q}\left[\frac{\partial R^c_{\psi}}{\partial \phi}\right],
\end{align}
where we define the critic-aware distribution $Q$ as 
\begin{align}
\label{eq:critic-aware-Q}
    Q(\epsilon) \propto \exp(\underbrace{
    -(R^c_{\psi} - R)^2/\tau}_{\text{Critic-awareness}}
    )\exp(\underbrace{-\frac{\epsilon^2}{2\sigma^2}}_{\mathclap{\text{Locality}}}).
\end{align}
This expectation allows us to incorporate the noisy, non-uniform nature of the critic's approximation of the objective by up-weighting the gradient computed at a point with a higher critic quality and down-weighting the gradient computed at a point with a lower critic quality. The first term in $Q$ reflects this, while the second term ensures that our estimation is based on a small region around the state-action sequence generated by the current, noise-free actor $\pi$. 

Since it is intractable to compute Eq.~\eqref{eq:critic-aware} exactly, we resort to importance sampling with the proposed distribution equal to the second term in Eq.~\eqref{eq:critic-aware-Q}. Then, our gradient estimate for the actor becomes the sum of the gradients from multiple realizations of the noisy actor in Eq.~\eqref{eq:noisy_actor}, where each gradient is weighted by the quality of the critic $\exp(-(R^c_{\phi} - R)^2 /\tau)$. $\tau$ is a hyperparameter that controls the smoothness of the weights. We observed in our preliminary experiment that the use of this critic-aware actor learning significantly stabilizes general learning of both the actor and critic.

\paragraph{Reference Translations for Training the Critic}~
In our setting of neural machine translation, we have access to a reference translation for each source sentence $X$, unlike in a usual setting of reinforcement learning. By force-feeding the reference translation into the underlying neural machine translation system (rather than feeding the decoded symbols), we can generate the reference state-action sequence. This sequence is much less correlated with those sequences generated by the actor, and facilitates computing a better estimate of the gradient w.r.t. the critic. 

%\paragraph{Break the Dependency} To avoid both the critic and the actor crash into a dangerous loop, it helps to keep the correct direction of the critic by providing the ground-truth trajectories to train the critic, or so-called \textit{teacher forcing}. More precisely, we directly feed the ground-truth $\hat{Y}$ into $\theta$ and generate a teacher sequence of hidden states $\hat{z}_{1:T}$. 

%\todo{this should go into the experimental details: Moreover, we shuffle the training set twice for separately training the critic and the actor. In this way, we can disconnect the dependency between consecutive updates and avoid a potential loop.}

% \paragraph{Train \bm{$\psi$} More} 

% As suggested in Section~\ref{sec.learn}, the critic is recommended to be sufficiently trained after each updates of the actor. %so that the critic can reasonably compute the gradient w.r.t. the actor in order to improve the true reward. 
% More precisely, we alternate training the critic and the actor with $N_c$ and $N_a$ steps respectively and keep $N_c > N_a$. 
 
In Alg.~\ref{algo2}, we present the complete algorithm. To make the description less cluttered, we only show the version of minibatch size = 1 which can be naturally extended. We also illustrate the proposed trainable greedy decoding and the proposed learning strategy in Fig.~\ref{fig:tgd}.

\section{Experimental Settings}

\begin{figure}[t]
\centering
\includegraphics[width=\linewidth]{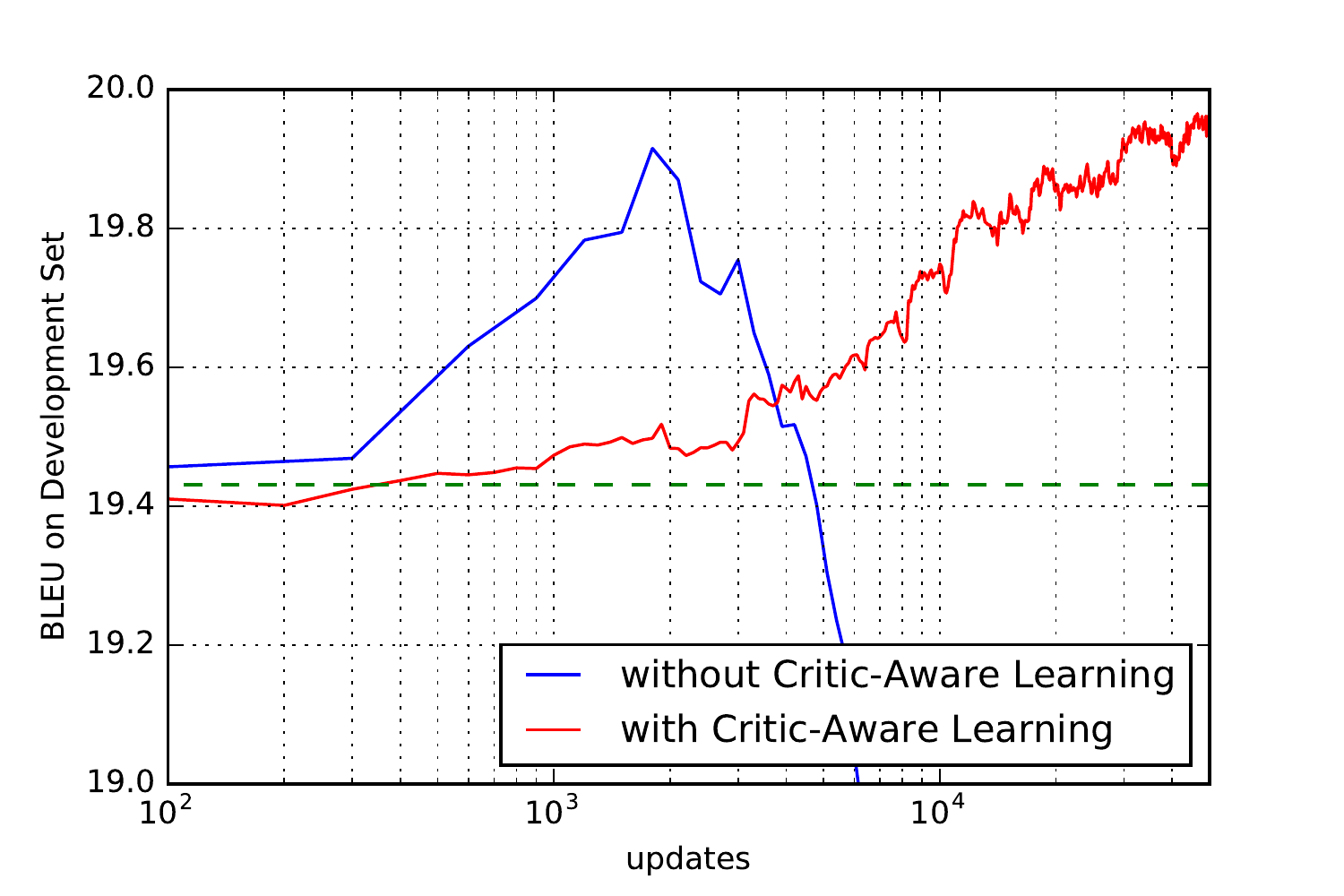}
\vspace{-4mm}
\caption{\label{fig:lr}  \small
Comparison of greedy BLEU scores whether using the critic-aware exploration or not on Ru-En Dataset. The green line means the BLEU score achieved by greedy decoding from the underlying NMT model.} %{\color{red} \bf KC: Jiatao, replace the y-axis label with "BLEU on Development Set"}}
\vspace{-4mm}
\end{figure}

We empirically evaluate the proposed trainable greedy decoding on four language pairs -- En-De, En-Ru, En-Cs and En-Fi -- using a standard attention-based neural machine translation system \citep{bahdanau2014neural}. We train underlying neural translation systems using the parallel corpora made available from WMT'15.\footnote{http://www.statmt.org/wmt15/} The same set of corpora are used for trainable greedy decoding as well. All the corpora are tokenized and segmented into subword symbols using byte-pair encoding (BPE) \citep{sennrich2015neural}. We use sentences of length up to 50 subword symbols for MLE training and 200 symbols for trainable decoding. For validation and testing, we use newstest-2013 and newstest-2015, respectively.

\subsection{Model Architectures and Learning}

\paragraph{Underlying NMT Model} 
For each language pair, we implement an attention-based neural machine translation model whose encoder and decoder recurrent networks have 1,028 gated recurrent units \citep[GRU,][]{cho2014learning} each. Source and target symbols are projected into 512-dimensional embedding vectors. We trained each model for approximately 1.5 weeks using Adadelta~\citep{zeiler2012adadelta}.

\vspace{-5pt}
\paragraph{Actor $\pi$}
We use a feedforward network with a single hidden layer as the actor. The input is a 2,056-dimensional vector which is the concatenation of the decoder hidden state and the time-dependent context vector from the attention mechanism, and it outputs a 1,028-dimensional action vector for the decoder. We use 32 units for the hidden layer with $\tanh$ activations.
%\todo{what kind of activation function? A: with tanh activation} 
 
\vspace{-5pt}
\paragraph{Critic $R^c$}

The critic is implemented as a variant of an attention-based neural machine translation model that takes a reference translation as a source sentence and a state-action sequence from the actor as a target sentence. Both the size of GRU units and embedding vectors are the same with the underlying model.  Unlike a usual neural machine translation system, the critic does not language-model the target sentence but simply outputs a scalar value to predict the true return. When we predict a bounded return, such as sentence BLEU, we use a sigmoid activation at the output. For other unbounded return like perplexity, we use a linear activation.
%\todo{what's the size of each critic? The critic has the same with the underlying model}
\vspace{-5pt}
\paragraph{Learning}

We train the actor and critic simultaneously by alternating between updating the actor and critic. As the quality of the critic's approximation of the decoding objective has direct influence on the actor's learning, we make ten updates to the critic before each time we update the actor once. We use RMSProp~\citep{tieleman2012lecture} with the initial learning rates of $2\times 10^{-6}$ and $2\times 10^{-4}$, respectively, for the actor and critic. 

We monitor the progress of learning by measuring the decoding objective on the validation set. After training, we pick the actor that results in the best decoding objective on the validation set, and test it on the test set.

\begin{figure*}[t]
\centering
\begin{minipage}{0.8\textwidth}
\includegraphics[width=\columnwidth]{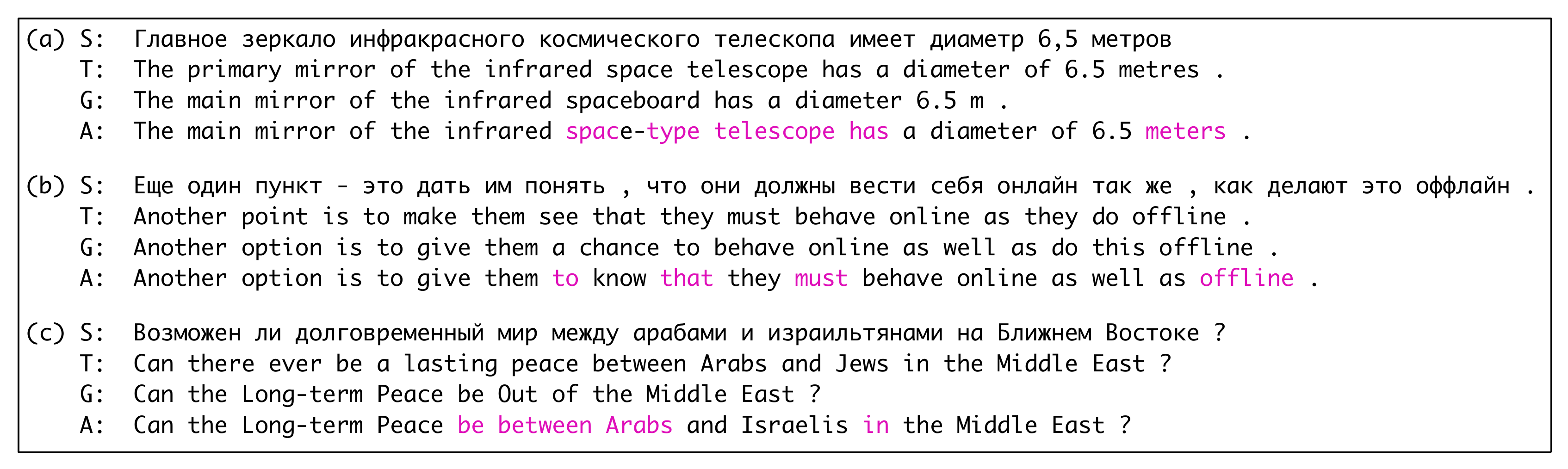}
\end{minipage}
\hfill
\begin{minipage}{0.19\textwidth}
\caption{\label{fig:exp}  \small Three Ru-En examples in which the difference between the trainable greedy decoding (A) and the conventional greedy decoding (G) is large. Each step is marked with magenta, when the actor significantly influenced the output distribution.} 
\end{minipage}

\vspace{-4mm}
\end{figure*}

\vspace{-5pt}
\paragraph{Decoding Objectives}

For each neural machine translation model, pretrained using maximum likelihood criterion, we train two trainable greedy decoding actors. One actor is trained to maximize BLEU (or its smoothed version for sentence-level scoring \citep{lin2004automatic}) as its decoding objective, and the other to minimize perplexity (or equivalently the negative log-probability normalized by the length.) %\todo{are we running experiments for this? A: yes, the experiments also finished.} 
%In addition to these two, we run a limited set of experiments in which the decoding objective includes a target translation length as well as the translation quality.

We have chosen the first two decoding objectives for two purposes. First, we demonstrate that it is possible to build multiple trainable decoders with a single underlying model trained using maximum likelihood learning. Second, the comparison between these two objectives provides a glimpse into the relationship between BLEU (the most widely used automatic metric for evaluating translation systems) and log-likelihood (the most widely used learning criterion for neural machine translation).% \todo{remove if no experiment of length control} 
%The last objective is included as an example of an arbitrary objective other than translation quality.
\vspace{-5pt}
\paragraph{Evaluation}
We test the trainable greedy decoder with both greedy decoding and beam search. Although our decoder is always trained with greedy decoding, beam search in practice can be used together with the actor of the trainable greedy decoder. Beam search is expected to work better especially when our training of the trainable greedy decoder is unlikely to be optimal. In both cases, we report both the perplexity and BLEU.

\subsection{Results and Analysis}

We present the improvements of BLEU and perplexity (or its negation) in Fig.~\ref{fig:result1} for all the language pair-directions. It is clear from these plots that the best result is achieved when the trainable greedy decoder was trained to maximize the target decoding objective. When the decoder was trained to maximize sentence-level BLEU, we see the improvement in BLEU but often the degradation in the perplexity (see the left plots in Fig.~\ref{fig:result1}.) On the other hand, when the actor was trained to minimize the perplexity, we only see the improvement in perplexity (see the right plots in Fig.~\ref{fig:result1}.) This confirms our earlier claim that it is necessary and desirable to tune for the target decoding objective regardless of what the underlying translation system was trained for, and strongly supports the proposed idea of trainable decoding.

The improvement from using the proposed trainable greedy decoding is smaller when used together with beam search, as seen in Fig.~\ref{fig:result1}~(b). However, we still observe statistically significant improvement in terms of BLEU (marked with red stars.) This suggests a future direction in which we extend the proposed trainable greedy decoding to directly incorporate beam search into its training procedure to further improve the translation quality. 

It is worthwhile to note that we achieved all of these improvements with negligible computational overhead. This is due to the fact that our actor is a very small, shallow neural network, and that the more complicated critic is thrown away after training. We suspect the effectiveness of such a small actor is due to the well-structured hidden state space of the underlying neural machine translation model which was trained with a large amount of parallel corpus. We believe this favourable computational complexity makes the proposed method suitable for production-grade neural machine translation \citep{wu2016google,crego2016systran}.

\paragraph{Importance of Critic-Aware Actor Learning}
In Fig.~\ref{fig:lr}, we show sample learning curves with and without the proposed critic-aware actor learning. Both curves were from the models trained under the same condition. Despite a slower start in the early stage of learning, we see that the critic-aware actor learning has greatly stabilized the learning progress. We emphasize that we would not have been able to train all these 16 actors without the proposed critic-aware actor learning.

\paragraph{Examples} 
In Fig.~\ref{fig:exp}, we present three examples from Ru-En. We defined the influence as the KL divergence between the conditional distributions without the trainable greedy decoding and with the trainable greedy decoding, assuming the fixed previous hidden state and target symbol. We colored a target word with magenta, when the influence of the trainable greedy decoding is large ($> 0.001$).  Manual inspection of these examples as well as others has revealed that the trainable greedy decoder focuses on fixing prepositions and removing any unnecessary symbol generation. More in-depth analysis is however left as future work. 

\section{Conclusion}

We proposed trainable greedy decoding as a way to learn a decoding algorithm for neural machine translation with an arbitrary decoding objective. The proposed trainable greedy decoder observes and manipulates the hidden state of a trained neural translation system, and is trained by a novel variant of deterministic policy gradient, called critic-aware actor learning. Our extensive experiments on eight language pair-directions and two objectives confirmed its validity and usefulness. The proposed trainable greedy decoding is a generic idea that can be applied to any recurrent language modeling, and we anticipate future research both on the fundamentals of the trainable decoding as well as on the applications to more diverse tasks such as image caption generating and dialogue modeling.

\section*{Acknowledgement}
KC thanks the support by Facebook, Google (Google Faculty Award 2016) and NVidia (GPU Center of Excellence 2015-2016). 
We sincerely thank Martin Arjovsky, Zihang Dai, Graham Neubig, Pengcheng Yin and Chunting Zhou for helpful discussions and insightful feedbacks.

%This work was partly supported by Samsung Electronics (Project: "Development and Application of Larger-Context Neural Machine Translation").

%\bibliography{acl2017}
\bibliography{acl2017}
\bibliographystyle{acl_natbib}

\end{document}